%% file: main.tex
\PassOptionsToPackage{capitalize,nameinlink,noabbrev}{cleveref}

\documentclass[10pt,twocolumn,letterpaper]{article}

\usepackage{cvpr}              % To produce the CAMERA-READY version

\definecolor{cvprblue}{rgb}{0.21,0.49,0.74}
\usepackage[pagebackref,breaklinks,colorlinks,allcolors=cvprblue]{hyperref}

\usepackage{graphicx}
\usepackage{tikz}
\usetikzlibrary{calc,positioning}

\newcommand\blfootnote[1]{%
  \begingroup
  \renewcommand\thefootnote{}\footnote{#1}%
  \addtocounter{footnote}{-1}%
  \endgroup
}
\usepackage{cleveref}

\newcommand{\figref}[1]{\figurename~\ref{#1}}

\def\paperID{*****} % *** Enter the Paper ID here
\def\confName{CVPR}
\def\confYear{2025}

\title{Enriched Feature Representation and Motion Prediction Module \\ for MOSEv2 Track of 7th LSVOS Challenge: 3rd Place Solution}

\author{Chang Soo Lim\textsuperscript{*}, Joonyoung Moon\textsuperscript{*}, Donghyeon Cho\textsuperscript{$\dagger$}\\
Computer Vision Lab., Department of Computer Science, Hanyang University\\
Seoul, South Korea\\
{\tt\small {\tt\small \{limjduni,joy999871,doncho\}@hanyang.ac.kr}}
}

\begin{document}
\maketitle
\blfootnote{*Equal contribution. $\dagger$ Corresponding author.}

\input{sec/0_abstract}    
\input{sec/1_intro}
\input{sec/2_method}
\input{sec/3_experiment}
\input{sec/4_conclusion}
{
    \small
    \bibliographystyle{ieeenat_fullname}
    \bibliography{main}
}


\end{document}

%% file: sec/0_abstract.tex
\begin{abstract}
Video object segmentation (VOS) is a challenging task with wide applications such as video editing and autonomous driving. 
While Cutie provides strong query-based segmentation and SAM2 offers enriched representations via a pretrained ViT encoder, each has limitations in feature capacity and temporal modeling. 
In this report, we propose a framework that integrates their complementary strengths by replacing the encoder of Cutie with the ViT encoder of SAM2 and introducing a motion prediction module for temporal stability. 
We further adopt an ensemble strategy combining Cutie, SAM2, and our variant, achieving 3rd place in the MOSEv2 track of the 7th LSVOS Challenge. We refer to our final model as \textbf{SCOPE (SAM2-CUTIE Object Prediction Ensemble)}. 
This demonstrates the effectiveness of enriched feature representation and motion prediction for robust video object segmentation. 
The code is available at \url{https://github.com/2025-LSVOS-3rd-place/MOSEv2_3rd_place}.
\end{abstract}

%% file: sec/1_intro.tex
\section{Introduction}
\label{sec:intro}
Video Object Segmentation (VOS)~\cite{caelles2017one, pont20172017, caelles20182018, ding2024lsvos} aims to segment target objects across all frames of a video, provided with an annotated mask in the first frame as supervision. 
Nowadays, with the widespread availability of video data, the VOS task has become more important than ever. 
It plays a crucial role in various applications such as autonomous driving~\cite{mei2022waymo}, video editing~\cite{cheng2024putting}, and augmented reality. 
Nevertheless, inherent challenges in VOS, including occlusion, object reappearance, and highly dynamic scenes, pose substantial difficulties for accurately estimating segmentation masks. 

To address these challenges, many research efforts have been devoted to advancing VOS. 
Recent query-based approaches such as Cutie~\cite{cheng2024putting} represent each target by a compact object vector, which is updated over time to maintain identity and robustness against occlusion and reappearance. 
However, Cutie relies on a ResNet~\cite{he2016deep}-based image encoder, which limits its ability to capture rich visual representations; leading to performance constraint when dealing with complex or long-term videos.

On the other hand, SAM2~\cite{ravi2024sam} demonstrates strong performance in VOS by utilizing a Hiera~\cite{ryali2023hiera}-based Vision Transformer encoder and memory attention mechanisms.
%
%SAM2~\cite{SAM2}, on the other hand, achieves strong performance in VOS by leveraging a powerful Hiera-based Vision Transformer encoder image encoder together with memory attention mechanisms. 
%
This pre-trained Hiera backbone in SAM2 extracts semantically rich and robust multi-scale features that generalize well across diverse video scenarios. 
While SAM2 demonstrates strong segmentation performance, it lacks explicit object tracking mechanisms, leading it unable to guarantee identity consistency in multi-object or long-occlusion scenarios.

\begin{figure*}[!t]
  \centering
  \begin{tikzpicture}
    \node[inner sep=0pt] (img) {\includegraphics[width=\textwidth]{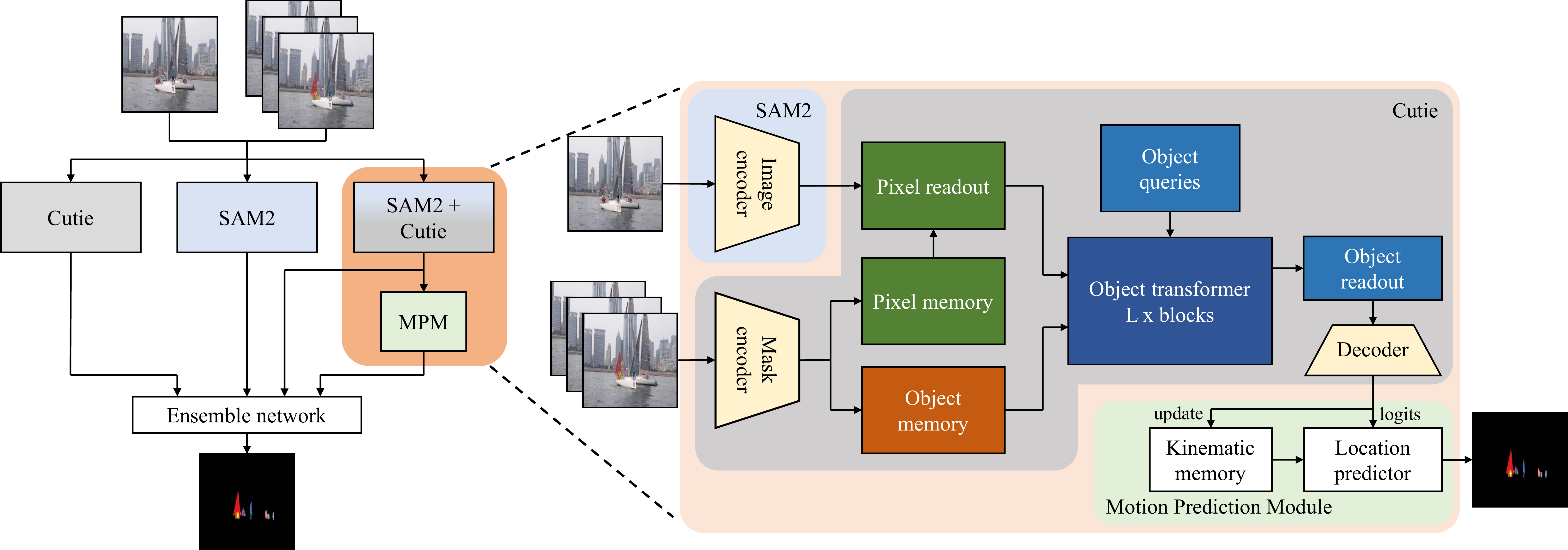}};
    \node[anchor=north west, font=\bfseries] at (img.north west) {(a)};
    \node[anchor=north, font=\bfseries] at ($(img.north west)!0.37!(img.north east)$) {(b)};
  \end{tikzpicture}
  \caption{\textbf{The overall framework of SCOPE.} The left figure (a) illustrates our overall ensemble pipeline, while the right figure (b) shows the fusion network of SAM2 and Cutie with the proposed Motion Prediction Module (MPM).}
  \label{fig:combined}
\end{figure*}

These difficulties are further magnified in Large-Scale Video Object Segmentation (LSVOS)~\cite{ding2024lsvos} challenge. 
LSVOS challenge succeeds on previous LVOS~\cite{hong2024lvos} benchmarks and extends them to more challenging datasets, aiming to evaluate VOS methods under longer sequences and more complex scenarios. 
This year, the 7th LSVOS Challenge consists of three tracks. 
The first is the Referring Video Object Segmentation (RVOS) track, which focuses on segmenting objects in videos based on natural language descriptions. 
The second is the classic VOS track, which evaluates methods on the MOSE~\cite{ding2023mose} dataset under a semi-supervised setting. 
Finally, the Complex Video Object Segmentation track introduces MOSEv2~\cite{ding2025mosev2}, a newly released dataset designed to assess segmentation performance in more challenging and realistic videos.

In this work, we focus mainly on the Complex Video Object Segmentation track of the 7th LSVOS Challenge, which uses the recently released MOSEv2 dataset. 
To address this complex dataset, our aim is to maximize the advantage of both SAM2 and Cutie by combining their complementary strengths.
Specifically, we replace the ResNet-based encoder of Cutie with the MAE pre-trained Hiera encoder from SAM2. 
In addition, since MOSEv2 contains frequent occlusions and object reappearances that often hinder reliable tracking, we propose a Motion Prediction Module (MPM) that predicts object positions during occlusion to enhance temporal consistency. 
Finally, to fully exploit the complementary strengths of both models, we design an ensemble pipeline that aggregates the logits of SAM2, Cutie, and our fused model using a learnable network. We denote this overall framework as \textbf{SCOPE (SAM2-CUTIE Object Prediction Ensemble)}.
The details of each component are described in the following sections.

%% file: sec/2_method.tex
\section{Method}
\label{sec:method}
As illustrated in~\figref{fig:combined}-(b), SCOPE is built on Cutie, where the original query encoder is replaced with the SAM2 image encoder to enrich semantic features~(\Cref{sec:SAM2}). 
We introduce an MPM that estimates the position of objects under occlusion to improve temporal consistency~(\Cref{sec:MPM}). 
Finally, as shown in~\figref{fig:combined}-(a), we design an ensemble strategy that integrates Cutie, SAM2, and our variant for further performance gains~(\Cref{sec:Ensemble}).
%
%As illustrated in \figref{fig:overview}, "our model" is built upon the Cutie architecture, where the Cutie query encoder is replaced with the SAM2 image encoder and augmented with an optional motion prediction module.
%
%As shown in \figref{fig:ensemble_network}, we also design an ensemble framework that integrates SAM2, Cutie, "our model" (motion prediction module enabled) and "our model" (motion prediction module disabled).
%
%The combined structure of SAM2 and Cutie is described in Section 2.1.
%
%The motion prediction module is introduced in Section 2.2, while the ensemble network is presented in Section 2.3.

%-------------------------------------------------------------------------
\subsection{Enriching Feature Representation Using SAM2}
\label{sec:SAM2}
Cutie, as mentioned above, leverages object vectors that enable consistent object tracking. 
However, due to its lightweight ResNet-based image encoder, the model struggles to capture rich representations, leading to degraded segmentation performance in long-term or complex videos. 
To enrich the feature representation in Cutie, we replaced its ResNet-based encoder with the MAE pre-trained Hiera image encoder from SAM2, which provides semantically rich and robust features. 
However, the Cutie encoder and the SAM2 image encoder produce different representations in both size and distribution, requiring semantic and dimensional alignment. 
To address this, we employ a $1{\times}1$ convolutional projection layer, through which the expressive image features of SAM2 can be effectively aligned and integrated into the tracking-oriented architecture of Cutie.
%
%However, the Cutie encoder and the SAM2 image encoder produce different representations, which means that both semantic and dimensional alignment are required. 
%
%To address this issue, we employ a $1{\times}1$ convolutional projection layer. 
%
%In this way, SAM2's expressive image features can be effectively aligned and utilized within Cutie's tracking-oriented architecture.
%-------------------------------------------------------------------------
\subsection{Motion Prediction Module}
\label{sec:MPM}
The model introduced in~\Cref{sec:SAM2}, which integrates the Cutie encoder and the SAM2 encoder, performs well in standard tracking cases. 
However, on challenging datasets such as MOSEv2, it often struggles when the target object temporarily disappears due to occlusion or leaving the field of view, or when multiple visually similar instances co-occur. 
To address these issues, we introduce an MPM that maintains an object-specific kinematic state (location, size, and velocity) of the target from recent frames and predicts the object position in the current frame under occlusion.
Based on this prediction, the MPM generates a Gaussian map centered at the predicted object position, which serves as a spatial prior for tracking. 
This map is combined with the segmentation logits of the VOS model via a weighted sum, guiding the model to focus on the most plausible region. 
By injecting the Gaussian map as a location-aware prior, MPM improves robustness to short-term disappearances and reduces confusion among similar objects, while remaining lightweight and optional when the prediction confidence is low.
%
%The integrated model of~\Cref{sec:SAM2} performs strongly in standard cases.
%
%However, on complex datasets such as MOSEv2, it often struggles when the target object temporarily disappears due to occlusion or leaving the field of view, or when multiple visually similar instances co-occur.
%
%To mitigate these failure modes, we introduce a Motion Prediction Module (MPM).
%
%MPM maintains an object-specific kinematic state (location, size, and velocity) estimated from recent frames and predicts the object’s location in the current frame.
%
%Based on this prediction, the module generates a Gaussian map centered at the estimated location, encoding the spatial prior of the object.
%
%This Gaussian prior is then combined with the segmentation logits of the base model through a weighted sum, effectively guiding the model to focus on the most plausible region.
%
%By injecting this location-aware prior, MPM enhances robustness against short-term disappearances and reduces confusion among similar objects while remaining lightweight and optional when the prediction confidence is low.

To this end, we continuously estimate the location, size, and velocity of each target object.
For initialization, given the binary mask $M_l \in \{0,1\}^{H \times W}$ of object $l$ in the first frame, the centroid $(\tilde{x}_0^l, \tilde{y}_0^l)$ and size $(\tilde{w}_0^l, \tilde{h}_0^l)$ are computed in pixels, normalized by the image resolution $(H,W)$ to form the relative state vectors as follows:
\begin{equation}
\mathbf{x}_0^l = \left(\frac{\tilde{x}_0^l}{W}, \frac{\tilde{y}_0^l}{H}\right), 
\qquad
\mathbf{u}_0^l = \left(\frac{\tilde{w}_0^l}{W}, \frac{\tilde{h}_0^l}{H}\right).
\label{eq:normalized_state}
\end{equation}
Also, the velocity $\mathbf{v}_0^l$ is set to zero vector.

At frame $t>0$, given the predicted mask $\hat{M}_t^l$ from the VOS model, the centroid and size are similarly computed and normalized to obtain $\hat{\mathbf{x}}_t^l$ and $\hat{\mathbf{u}}_t^l$. 
The state is then updated with an exponential moving average (EMA):
\begin{equation}
\mathbf{x}_t^l = \alpha \mathbf{x}_{t-1}^l + (1-\alpha)\hat{\mathbf{x}}_t^l,
\
\
\mathbf{u}_t^l = \alpha \mathbf{u}_{t-1}^l + (1-\alpha)\hat{\mathbf{u}}_t^l,
\label{eq:update_ema}
\end{equation}
where $\alpha\in(0,1)$ balances stability and responsiveness.
The velocity is defined as the displacement of consecutive centroids:
\begin{equation}
\mathbf{v}_t^l = \mathbf{x}_t^l - \mathbf{x}_{t-1}^l
\label{eq:update_velocity}
\end{equation}
without EMA to preserve sensitivity to sudden motion.
If no valid mask $\hat{M}_t^l$ is available, the location is extrapolated using the last known velocity while the object size remains unchanged.
Finally, to incorporate the estimated kinematics, we generate a Gaussian map over the image at each frame. 
For each pixel $(i,j)$, its value is defined as
%
%To incorporate spatial priors, we generate a Gaussian map over the image at every frame.
%
%For each pixel index $(i,j)$, the value is defined as
\begin{equation}
G_t^l(i,j) = \exp\left(
-\frac{(i/W - x_t^l)^2}{2\sigma_x^2}
-\frac{(j/H - y_t^l)^2}{2\sigma_y^2}
\right),
\label{eq:gaussian_map}
\end{equation}
where the center $(x_t^l,y_t^l)$ corresponds to the predicted object location and the variances $\sigma_x, \sigma_y$ are set proportional to the estimated width $w_t^l$ and height $h_t^l$.
This design adaptively scales the Gaussian distribution with the object size, yielding sharper priors for small objects and broader ones for large objects.
The Gaussian map is then integrated with the output of the segmentation network to bias the model toward the predicted region.
Specifically, let $\hat{Z}_t^l$ denote the raw logits of object $l$ in frame $t$.
We combine them with the Gaussian map through a weighted sum:
\begin{equation}
{Z}_t^l = \hat{Z}_t^l + \beta\cdot\log(G_t^l + \epsilon),
\label{eq:gaussian_weighted_sum}
\end{equation}
where $\beta$ controls the influence of the prior and $\epsilon$ is a small constant for numerical stability.
This formulation effectively increases the confidence of pixels near the predicted location while suppressing unlikely regions.
By applying this fusion to every frame, the module consistently injects location-aware information into the segmentation process.
As a result, the model can recover more gracefully from short-term disappearance (e.g., due to occlusion) and is less prone to confusion when multiple visually similar objects co-occur.
Importantly, MPM remains lightweight and optional: when the base network already produces confident predictions, the Gaussian prior has little influence, while in ambiguous cases it provides additional guidance to resolve uncertainty.
%-------------------------------------------------------------------------
\subsection{Ensemble Network}
\label{sec:Ensemble}
To further improve robustness and accuracy, we adopt an ensemble strategy that combines the complementary strengths of multiple models. 
Specifically, as shown in~\figref{fig:combined}-(a), we integrate four components: the original SAM2, the original Cutie, SAM2 + Cutie with MPM, and SAM2 + Cutie without MPM.
The MPM-off variant is included to preserve fine-grained details, as the Gaussian map, although beneficial under occlusion, tends to oversmooth boundaries.
%
%Specifically, as shown in~\figref{fig:combined}-(b), we integrate four components: SAM2, Cutie, "our model" (MPM on) and "our model" (MPM off).
%
%The inclusion of the MPM-off variant is motivated by the fact that the Gaussian prior, while useful under occlusion, may oversmooth boundaries; thus, the MPM-off model output preserves fine-grained details.
%
By combining all four models, the ensemble can retain the complementary advantages of each while reducing the impact of their individual weaknesses.

Formally, let $Z_C, Z_S, Z_{M-}, Z_{M+}$ denote the logits from Cutie, SAM2, SAM2 + Cutie without MPM, and SAM2 + Cutie with MPM, respectively, all aligned to the same spatial resolution $(H,W)$. 
These outputs are then fed into a shallow fusion module $f_\theta$:
\begin{equation}
F = f_{\theta}(Z_C, Z_S, Z_{M-}, Z_{M+})
   \in \mathbb{R}^{(N+1)\times H \times W},
\label{eq:ensemble_fusion}
\end{equation}
where $N$ is the number of object classes.
Note that~\eqref{eq:ensemble_fusion} computes a weighted combination and produces the final ensemble logits.
This design enables the ensemble to leverage the complementary strengths of all components while mitigating their individual weaknesses.

\if 0
Formally, let
\begin{equation}
\begin{array}{c}
Z_C \in \mathbb{R}^{(N+1)\times H \times W}, \\
Z_S \in \mathbb{R}^{N\times H \times W}, \\
Z_{M-} \in \mathbb{R}^{(N+1)\times H \times W}, \\
Z_{M+} \in \mathbb{R}^{(N+1)\times H \times W},
\end{array}
\label{eq:ensemble_inputs}
\end{equation}
denote the logits produced by Cutie, SAM2, SAM2 + Cutie without MPM, and SAM2 + Cutie with MPM, respectively.
%
Here $N$ is the number of object classes, and $H\times W$ is the shared spatial resolution (all outputs are bilinearly aligned with $(H,W)$ if needed).
%
These four model outputs are then fed into a shallow fusion module $f_\theta$:
\begin{equation}
F = f_{\theta}(Z_C, Z_S, Z_{M-}, Z_{M+})
   \in \mathbb{R}^{(N+1)\times H \times W},
\label{eq:ensemble_fusion}
\end{equation}
which computes a weighted combination of the four models and produces the final ensemble logits.
%
This design allows the ensemble to exploit the complementary strengths of all components while suppressing their individual weaknesses.
\fi

%% file: sec/3_experiment.tex
\section{Experiment}
\label{sec:experiment}
%-----------------------------------------------------
\subsection{Training}
\noindent \textbf{Segmentation Network.} 
The segmentation network was initialized with the pre-trained parameters of SAM2 and Cutie.
To mitigate the misalignment between their image encoders, a $1{\times}1$ convolutional projection layer was pre-trained with an L2 loss between the outputs of the two encoders, ensuring that the projected features were aligned with those of Cutie.
The network was first fine-tuned in the original MOSE dataset for 5 epochs, and the three checkpoints with the highest J\&F scores were combined using the model soups~\cite{wortsman2022model} method.
We then further trained the model on MOSE for 3 epochs using the soup weights, followed by 8 epochs of fine-tuning on MOSEv2.
All training was performed on two A100 GPUs with a batch size of 8, and the learning rate was adjusted each epoch based on the validation performance of the corresponding checkpoint.

\noindent \textbf{Motion Prediction Module (MPM).} 
The MPM was trained while freezing the segmentation network composed of SAM2 and Cutie.
It was optimized with a cross-entropy loss between the blended logits and the ground-truth masks, enabling the learned prior to sharpen object localization while maintaining temporal coherence.
Training was conducted using AdamW optimizer~\cite{loshchilov2019decoupled} ($\text{lr}=1{\times}10^{-4}$, weight decay $1{\times}10^{-6}$) with 1.0 gradient clipping.
To enhance stability, five gradient update steps were performed for each annotated frame, and the module parameters were re-initialized at the beginning of every video sequence.
An exponential moving average with momentum $\alpha=0.9$ was further applied to the states of objects across frames.

\begin{figure*}[t]
    \centering
    \includegraphics[page=1,width=\textwidth,keepaspectratio]{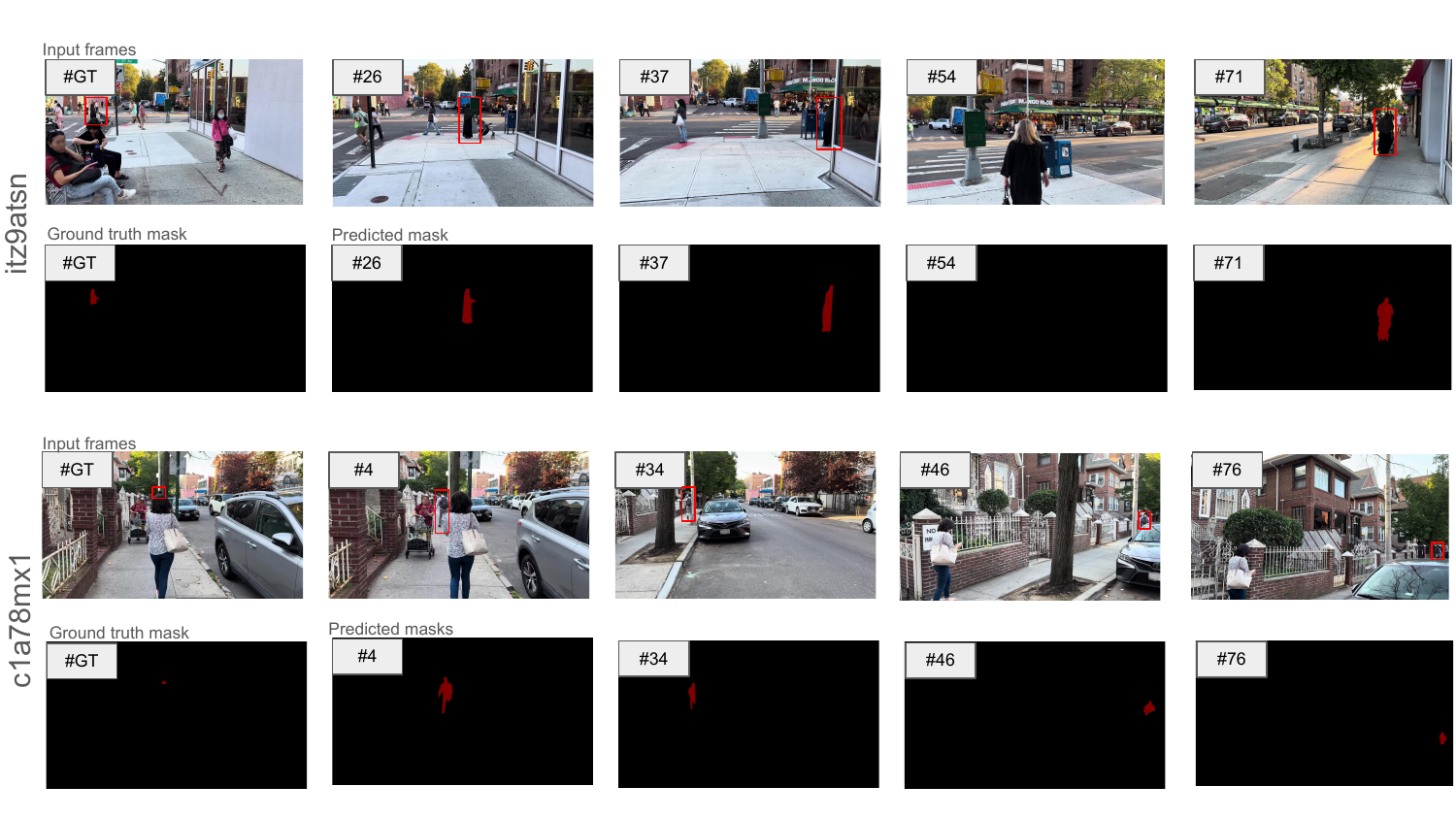}
    \caption{\textbf{Qualitative results on the MOSEv2 test set.} Each row shows the input frame and predicted mask, given the ground truth mask in the first frame. The target object is highlighted with a red bounding box in the input frames. As can be seen, SCOPE is able to robustly track the target across challenging scenarios such as occlusions, scale variations, and cluttered backgrounds.}
    \label{fig:mosev2test}
\end{figure*}

\begin{table}[t]
\centering
\begin{tabular}{lcccc}   
\hline
Rank & Participant & J\&F\raisebox{1.0ex}{\scalebox{0.8}{$\boldsymbol{\cdot}$}} & J & F\raisebox{1.0ex}{\scalebox{0.8}{$\boldsymbol{\cdot}$}} \\
\hline
1 & mmm & 39.89 & 39.02 & 40.76 \\
2 & qqqqaaaa & 39.70 & 38.87 & 40.53 \\
\textbf{3} & \textbf{limjduni} & \textbf{37.87} & \textbf{36.99} & \textbf{38.75} \\
4 & waaaaaaaaaa & 35.77 & 34.98 & 36.56 \\
5 & springggg & 35.39 & 34.63 & 36.15 \\
\hline
\end{tabular}
\caption{\textbf{MOSEv2 test benchmark.} J measures region similarity (Jaccard index), F\raisebox{1.0ex}{\scalebox{0.8}{$\boldsymbol{\cdot}$}} denotes the improved boundary accuracy (modified F-measure), and J\&F\raisebox{1.0ex}{\scalebox{0.8}{$\boldsymbol{\cdot}$}} denotes their average score.}
\label{tab:mosev2_results}   
\end{table}

\noindent \textbf{SAM2 \& Cutie.} 
For the ensemble network, SAM2 and Cutie are included as individual models initialized with their pre-trained weights.
To improve their performance on MOSEv2, both models were fine-tuned for 8 epochs while preserving their original initialization.

\noindent \textbf{Ensemble Network.} 
The ensemble network is trained to fuse logits from four different models.
In this network, each branch include learnable scalar weights for foreground and background channels, as well as bias terms and temperature parameters.
All weights are initialized to 1.0, all biases to 0.0, and the temperature parameters to 1.2.
To stabilize learning, the temperatures are transformed with a softplus function (with a $10^{-3}$ offset) and clamped to the range $[0.8, 2.0]$. 
The network is optimized with cross-entropy loss against ground-truth masks for 8 epochs on the MOSEv2 training set.
Optimization is performed using AdamW optimizer~\cite{loshchilov2019decoupled}~($\text{lr}=1{\times}10^{-5}$, weight decay $1{\times}10^{-4}$) with cosine decay scheduling and 200 warm-up steps, and the gradients are reduced to a maximum norm of 1.0.

%-----------------------------------------------------
\subsection{Inference}
For inference, the segmentation network combines SAM2 with Cutie, where the SAM2 component is run with its default configuration, while Cutie is configured with \texttt{top\_k}=60, \texttt{max\_mem\_frames}=12, an input image size of 960, and \texttt{mem\_every}=3, without using long-term memory.
All other modules and networks, including the MPM and the ensemble network, are used with the same configurations as during training.

\subsection{Main Results}
In the 7th LSVOS Challenge, our method ranked third overall and demonstrated competitive performance against most participants.
Table~\ref{tab:mosev2_results} presents the test results on MOSEv2, reporting the Jaccard value (J), the modified F-measure (F\raisebox{1.0ex}{\scalebox{0.8}{$\boldsymbol{\cdot}$}}), and their average (J\&F\raisebox{1.0ex}{\scalebox{0.8}{$\boldsymbol{\cdot}$}}).
Specifically, our method achieved a Jaccard value of 36.99, a modified F-measure of 38.75, and overall J\&F\raisebox{1.0ex}{\scalebox{0.8}{$\boldsymbol{\cdot}$}} score of 37.87.

Furthermore,~\figref{fig:mosev2test} illustrates the robustness of SCOPE in handling challenging scenarios, including small-object segmentation and dynamic video scenes with occlusions.
Since Cutie originally employs a ResNet-based query encoder, it may struggle to capture semantically rich features, which can limit its performance on small objects.
By replacing it with the SAM2 image encoder, we significantly enhanced the representational capacity of the model.
In addition, the MPM module further strengthens the model’s ability to capture diverse object appearances and maintain temporal consistency in complex scenarios.
Qualitative examples in~\figref{fig:mosev2test} highlight these improvements.
In the second row, SCOPE successfully re-identifies and tracks the target even when the object temporarily disappears from view and reappears later.
In the fourth row, our method maintains robust tracking performance when the object moves farther away and is observed from diverse camera angles.
% \subsection{Ablation Study}

%% file: sec/4_conclusion.tex
\section{Conclusion}
In this paper, we have exploited the advantages of two powerful Video Object Segmentation models through fusion and ensembling. 
To further enhance robustness on complex video scenarios such as MOSEv2, we have proposed a Motion Prediction Module (MPM) designed to handle challenges including occlusions, reappearance, and dynamic scenes. 
Our solution demonstrated its effectiveness and robustness in the 7th LSVOS Challenge, where it achieved the third place with a notable J\&F score of 37.87. 
This result highlights the capability of our method to robustly and effectively handle the highly complex video scenarios in video object segmentation tasks.
\label{sec:conclusion}

\noindent \paragraph{Acknowledgements.} This work was supported by the National Research Foundation of Korea(NRF) grant funded by the Korea government(MSIT) (RS-2025-00521432).